\documentclass{llncs}
\usepackage{mathptmx}
\usepackage{helvet}
\usepackage{courier}
\usepackage{makeidx}
\usepackage{graphicx}
\usepackage{multicol}
\usepackage[bottom]{footmisc}

\usepackage[utf8]{inputenc}
\usepackage{hyperref}

\begin{document}
\title{Leaf Identification Using a Deep Convolutional Neural Network}
\author{Christoph Wick \and Frank Puppe}
\institute{University of W\"urzburg, Am Hubland, 97074 W\"urzburg, Germany, \\
	\email{christoph.wick@uni-wuerzburg.de}}

\maketitle

\begin{abstract}
Convolutional neural networks (CNNs) have become popular especially in computer vision in the last few years because they achieved outstanding performance on different tasks, such as image classifications.
We propose a nine-layer CNN for leaf identification using the famous Flavia and Foliage datasets.
Usually the supervised learning of deep CNNs requires huge datasets for training. However, the used datasets contain only a few examples per plant species.
Therefore, we apply data augmentation and transfer learning 
to prevent our network from overfitting.
The trained CNNs achieve recognition rates above 99\% on the Flavia and Foliage datasets, and slightly outperform current methods for leaf classification.
\end{abstract}

%===============================================
\section{Introduction}
%===============================================

Currently, supervised learning of convolutional neural networks (CNNs) for classification tasks achieve state-of-the-art performances on a wide range of datasets, e.g. MNIST \cite{mnistlecun} and ImageNet \cite{imagenet_cvpr09}.
Even though these datasets are usually huge in the amount of examples per class optimum values are mostly achieved by using data augmentations.
The Flavia \cite{Wu:2007} and Foliage \cite{Kadir:2011} datasets used in this paper include approximately 60 images per class in Flavia and 120 in Foliage, which is why data augmentation is extremely important to obtain a reliable generalization of the trained networks.
Moreover, we apply further techniques that help prevent overfitting of the network.
Those are dropout \cite{Srivastava:2014} and transfer learning which provides an initial guess for the weights of the network, e.g. \cite{Donahue:2013}.
As a result, the trained 9-layer CNNs achieve outstanding recognition rates above 99\% that also outperform slightly the current state-of-the-art published by Sulc and Matas \cite{Sulc:2015}, who utilize a texture-based leaf identification based on local feature histograms.

The following paper is structured as follows. At first, we introduce similar work which is also used to compare our results. Next, we present the model we used. This includes preprocessing, the network structure, batch generation, data augmentation, pretraining, and the execution parameters of our experiments.
Finally, we demonstrate the influence of augmentations, pretraining, and dropout on the accuracy and compare our results to other published values.

%===============================================
\section{Related Work}
%===============================================

One of the first important datasets for leaf classification is the Flavia dataset that was introduced by Wu et al. \cite{Wu:2007}. The applied probabilistic neural networks obtained accuracies around 90\% using 12 simple geometric features.

Kadir et al. \cite{Kadir:2011} applied several different features on the task of image classification on the Flavia dataset, but also on their initially published Foliage dataset.
Their best methods are based on probabilistic neuronal networks which use features derived from a leaf's shape, vein, color, and texture.
These models achieve accuracies of about 95\% on the Flavia and 95.75\% on the Foliage dataset \cite{Kadir:2012}.

A recent publication of Sulc and Matas \cite{Sulc:2015} use a rotation and scale invariant version of local binary patterns applied to the leaf interior and to the leaf margin. A Support Vector Machine classification yields impressive state-of-the art recognition values of above 99.5\% on the Flavia dataset and 99.0\% on the Foliage dataset.

Reul et al. \cite{Reul:2016} use a 1-nearest-neighbor classifier based on contour, curvature, color, Hu, HOCS, and binary pattern features. They achieve accuracies of about 99.37\% on the Flavia dataset and 95.83\% on the Foliage dataset.

A similar approach to our work that is also based on deep CNNs was published by Zhang et al. \cite{Zhang:2015} who use a 7-layer CNN and a comparable data augmentation for leaf classification and achieved results of approximately 95\% on the Flavia dataset.
Our work differs mainly by usage of arbitrary rotations, of a larger 9-layer CNN and of a pretrained model that initializes the network weights. Moreover we generate new augmentations during the training which will result in so to say infinite different images, while Zhang et al. augment the dataset by a given factor before training.
This procedure will be explained in Sects. \ref{sec:batch_generation} and \ref{sec:dataset_augmentation}.

Transfer learning across datasets by learning features on large-scale data  and then transferring them to a different tasks has proven to be successful e.g. in \cite{Donahue:2013} or \cite{Li:2010}.
We implement a ``supervised pretraining'' approach, i.e. we train a network on the Caltech-256 dataset \cite{Griffin:2007} and use the resulting network weights as initial conditions for all further trainings.
Therefore each following training task does not start with randomly initialized weights but instead with ones that already represent probably useful features. 

Dataset augmentation is one of the key concepts for learning deep convolutional neural networks due to the power of those deep networks to generalize on large datasets. For example all of the state-of-the-art accuracies \cite{Ciresan:2010}\cite{Ciresan:2011}\cite{Meier:2011} on the famous MNIST dataset \cite{mnistlecun}  make use of random distortions to augment the dataset.
We make use of linear label-preserving transformations that are presented in Sect. \ref{sec:dataset_augmentation}.

%===============================================
\section{Models}\label{sec:models}
%===============================================

In the following we will introduce the preprocessing of each dataset, the network structure, the generation of batches, the data augmentation, the pretraining, and finally the parameters during the training.

%===============================================
\subsection{Preprocessing}\label{sec:preprocessing}
%===============================================

Each of the images in the used datasets is preprocessed in order to generate standardized input data for the training process of the network.
A segmentation step is not required because all leaves in the Foliage or Flavia dataset are already photographed on a white background and had their petioles removed.

For preprocessing we first compute the bounding box of a leaf to determine the relevant data, extract the enclosed image, and resize it to $344\times 344$ pixels.
By adding a white margin of $3$ pixels on each side we finally end up with a $350\times 350$ pixels sized image.

Before training a single network we split the complete dataset $D$ containing $C$ classes and $N$ images into a training set $S_{TR}$ and a testing set $S_{TE}$ according to the ratio $N_{TE}(C)/N_{TR}(C)$, where $N(C)$ defines the number of images per class.
For example the notation $10\times 40$ states that 10 images out of $N$ are randomly moved to the testing set and 40 images out of the remaining ones are randomly inserted in the training set. Therefore the total number of instances in the testing set is $10 \times C$ and the total number of instances in the training set is $40 \times C$. Note that in this example possibly not all examples of $D$ are used. The notation $10\times \textsc{All}$ means that $10$ instances per class are moved to the training set and the rest is used for testing.

Note that the splitting of the Foliage dataset into $S_{TR}$ and $S_{TE}$ is prescribed, which we will denote  as $\textsc{Fixed}$.
Therefore, we apply the random distribution only to Flavia.

%===============================================
\subsection{Network structure}\label{sec:network_structure}
%===============================================

\begin{figure}[tb]
\centering
\includegraphics[width=\textwidth]{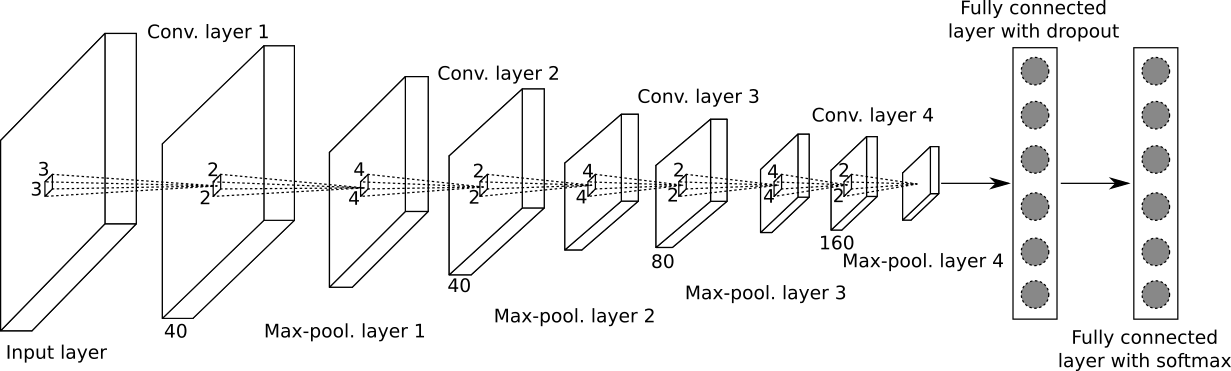}
\caption{
	The used CNN structure.
	The two numbers at the small block indicate the kernel sizes in convolutional or max-pooling layers.
	The digit below a convolutional layer specifies the number of kernels.
	The stride of convolutional and max-pooling layers it is 1 and 2, respectively.
	The first fully connected layer consists of 500 output nodes.}
\label{fig:network_structure}
\end{figure}

We use a deep CNN structure shown in Fig. \ref{fig:network_structure}  that consists of four convolutional layers (see e.g. \cite{LeCun:1998}), each one is followed by a max-pooling layer, a ReLU (see e.g. \cite{Krizhevsky:2012}) fully-connected layers, and a softmax layer indicating the class of the input image.
The size of the input layer is $300\times 300 \times 3$ pixels, whereby $3$ is the number of color channels.
Note that the size is smaller than the one used in the preprocessing to allow cropping as data augmentation, see Sect. \ref{sec:dataset_augmentation}.
The number of nodes in the output layer matches the count of classes $C$ of the used dataset.

%===============================================
\subsection{Batch generation}\label{sec:batch_generation}
%===============================================

Training is performed by utilizing batches of size 32 that are renewed before each iteration consisting of a forward and backward pass. A single element of a batch is created by choosing a random class $c \in C$ and afterwards a corresponding image $i \in N(c)$ out of the images in the training dataset.
Even if $N(c)$ is different for distinct $c$, e.g. in a $10\times \textsc{All}$ splitting, we thereby achieve a uniform class distribution in our training set.

%===============================================
\subsection{Dataset augmentation}\label{sec:dataset_augmentation}
%===============================================

Each randomly generated batch is altered by label-preserving transformations (e.g. \cite{Simard:2003}) to reduce overfitting.
A single transformation $T$ is performed by applying the following elementary operations on a single image in the batch:
\begin{itemize}
  \item Rotation: Rotating the image by an arbitrary angle.
  \item Scaling: Scaling the image by a factor in the range of $2^{[-0.1,0.1]}$.
  \item Cropping: Selection of a window with a size of $300 \times 300$ pixels out of the (transformed) image.
  \item Contrast: Multiplication of the color values by a factor of $2^{[-1, 1]}$.
  \item Brightness: Adding a value in the range of $[-20, 20]$ to the image colors.
  \item Flip: Mirroring the image window.
\end{itemize}

Note that changing the contrast and brightness also changes the white background of the images.

During training we use uniformly random distributed transformations $T_R$.
For testing we use $T_R$, the original image (no transformation) $T_0$ and a transformation $T_F$ that simply generates rotations of an original image by a fixed angle offset.

%===============================================
\subsection{Pretraining}
%===============================================

We use a ``supervised pretraining'' method also known as ``transfer learning'' similar to the one applied in \cite{Donahue:2013}.
The main idea is to use a large-scale dataset that differs substantially from the actual smaller dataset for a supervised pretraining phase.
The trained weights of all layers except the softmax-layer are then used as initial conditions for training the actual dataset.
Thus the filters of the convolutional layers are not initialized randomly but instead set to pretrained values that already learned useful generalizing features.

For that purpose we use the Caltech-256 dataset \cite{Griffin:2007} which consists of classes of images that are entirely different from leaf classifications, e.g. various animals, tools, objects, vehicles or even fictional characters.
This training is performed by using the same preprocessing as described in Sect. \ref{sec:preprocessing} and the same dataset augmentation as introduced in Sect. \ref{sec:dataset_augmentation}.

The increase of performance of the trained network on the Flavia and Foliage dataset shall be studied deeper in Sect. \ref{sec:results_dataaug}.
In our experiment we only trained once on the Caltech dataset and used these results as initial network weights.

%===============================================
\subsection{Experiments}
%===============================================

The CNNs implementation is based on the caffe framework \cite{Caffe:2014} running on a single NVIDIA Titan X GPU.
The calculation of new training batches is performed simultaniously on a i7-5820K CPU using OpenCV \cite{opencv:2000} and OpenMP \cite{openmp:1998}.

Each network is trained with a batch size of 32 using a Nesterov solver \cite{Netserov:1987} and a momentum of 0.95. 
The training lasts for exactly 50000 iterations.
During the process we multiply the initial learning rate of 0.001 by 0.1 each 20000 iterations.
Moreover, we use a L2 regularization with a weight decay of $0.0005$.
Training the network on the Caltech dataset to generate the pretrained weights takes approximately 8.5 hours, but using 100000 iterations.
Training of a single network afterwards took approximately 4.2 hours.

For each model we train 10 networks that use different random seeds for splitting the dataset into testing and training, generating the random batches and performing the data transformation.
Since the testing and training parts are prescribed in the Foliage dataset, a single run only changes the seed for generating a random batch and data transformations.

All stated accuracies and their errors are obtained by averaging the results of the 10 networks and computing the standard deviation.

%===============================================
\section{Results}
%===============================================

First, we will study the effect of the presented methods on the accuracy of the models.
Afterwards, we show the final evaluation of the best models and discuss the occurring errors.
During the classification, the predicted leaf class is chosen to be the one with the highest probability in the output softmax layer.

\subsection{Data augmentation}\label{sec:results_dataaug}

In the following we study the effect of a single method shown in Sect. \ref{sec:models} on the accuracy or error.
Therefore, we compare the result of a model that uses all augmentation methods to one that leaves out a single one.
The seeds in all of these modes are identical, i.e. the images in the testing and training sets are the same.
We use the Flavia dataset and a $10 \times 40$ splitting for evaluation.
A data point in the following plots is computed every 500 iterations by predicting 3200 augmented examples that are chosen randomly, using $T_R$ on the test dataset.
For smoothing the graph we average over 5 data points.
Note that the jump at iteration 20000 occurs due to a change of the learning rate at this point.

\begin{figure}[tb]
\centering
\includegraphics[width=1\textwidth]{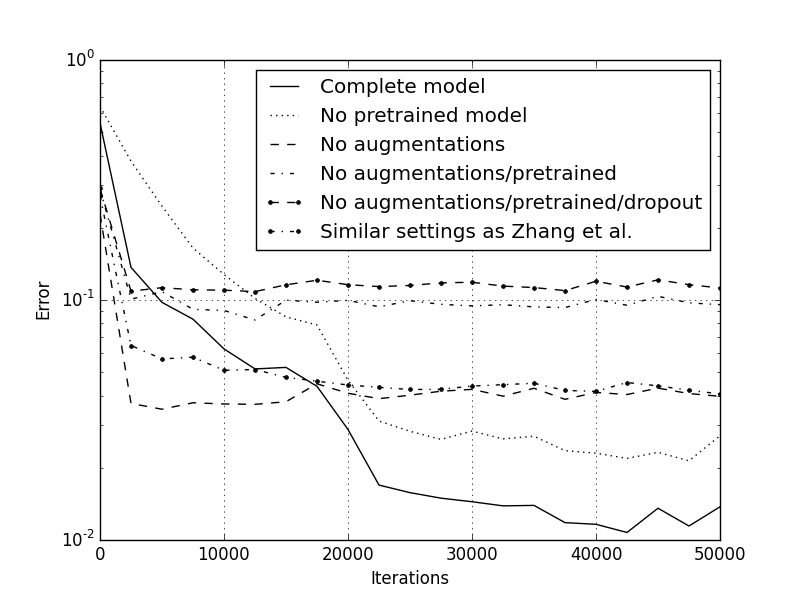}
\caption{The error on test examples is computed by a model with or without the use of pretraining.}
\label{fig:pretrained}
\end{figure}

The effect of pretraining and data augmentation is shown in Fig. \ref{fig:pretrained}.
The solid line indicates a typical learning process including all of the presented methods.
It is clearly visible that the usage of pretraining leads to a lower classification error, because a model learned without a pretraining (dotted line) shows a constant higher error.
The absence of data augmentation (dashed curve) leads to a initial fast learning followed by a constant plateau.
This behavior can be explained by the absence of data augmentations as rotation or color changes that drastically increase the number of different examples that have to be learned.
Moreover, in the dashed curve the effect of overfitting becomes visible.
The accuracy reaches a maximum at approximately 10000 iterations and then decreases due to overfitting the training data.
Usage of a model without a pretrained model or any data augmentation lead to a even worse result (dash dotted line).
In addition removing dropout that helps prevent overfitting deteriorates (dashed line with big dots).

The dash dotted line with big dots shows the results when training a model that is very similar to the settings of Zhang et. al \cite{Zhang:2015}.
For that we applied clipping with a maximum offset of $0.1$ of the allowed range, a scaling factor between $0.9$ and $1.1$, rotations by an angle up to 10 degrees, and a contrast factor between $0.8$ and $1.2$.
Changes in contrast are neglected.
It is clearly visible that the accuracy is coincidentally very similar to the model without any data augmentation.
The value of approximately $97\%$ is slightly higher than the reported one of $94\%$.
This difference can be explained by the larger overall structure of the network, as well as by the usage of a training set that repeatedly generates new random batches instead of using a fixed extended dataset as used by \cite{Zhang:2015}.
Furthermore, our result of about $89\%$ by usage of a conventional structure (dashed line with big dots) is very similar to $87\%$ reported by Zhang et al. \cite{Zhang:2015}.

\subsection{Final results}\label{sec:final_results}

%===============================================

For evaluating the networks we use the testing set $S_{TE}$.
Since every step during the learning process is probabilistic, reliable results require averaging of different models.
For this purpose, we take the mode (most occurring prediction) of an augmented version of each single image in the testing set, which is also called ``oversampling''.
For this purpose we generate 64 augmentations based on random operations $T_R$ and on fixed rotations $T_F$ with an offset of $64/360$ degrees.
As comparative value we denote the prediction of all testing images without any augmentation by $T_0$ .
To overcome randomness of the seeds for splitting the dataset or generating random numbers for data augmentation we average 10 runs.

Table \ref{tab:results} shows the classification results of our CNNs on the Flavia and Foliage dataset averaged over 10 runs compared to the results of similar publications.

\begin{table}[tb]
	\caption{Evaluation of the trained networks using the single image and augmentation of each image based on $T_R$ and $T_F$. All numbers are given as a percentage. The best values are marked bold.}
	\begin{tabular}{|l|c|c|c|c|}
	\hline
	 & Flavia & Flavia & Flavia & Foliage \\
	 & $10 \times \textsc{All}$ & $10 \times 40$ & $1/2 \times 1/2$ & \textsc{Fixed} \\
	\hline
	\hline
	Single image $T_0$ & $99.38 \pm 0.55$ & $99.41 \pm 0.43$ & $99.39 \pm 0.40$ & $98.77 \pm 0.39$ \\
	Av. $T_R$ & $99.72 \pm 0.31$ & $\bf 99.75 \pm 0.29$ & $\bf 99.67 \pm 0.20$ & $99.28 \pm 0.11$ \\
	Av. $T_F$ & $\bf 99.81 \pm 0.26$ & $99.69 \pm 0.33$ & $99.66 \pm 0.19$ & $\bf 99.40 \pm 0.09$ \\
	\hline
	\hline
	Sulc and Matas \cite{Sulc:2015} & $-$ & $99.7 \pm 0.3$ & $99.4 \pm 0.2$ & $99.0$ \\
	Reul et al. 
	\cite{Reul:2016} & $-$ & $99.37 \pm 0.08$ & $-$ & $95.83$ \\
	Kadir et al. \cite{Kadir:2012} & $-$ & $95.0$ & $-$ & $ 95.75 $ \\
	Zhang et al. \cite{Zhang:2015} & $94.69$ & $-$ & $-$ & $-$ \\
	\hline
	
	\end{tabular}
	\label{tab:results}
\end{table}

As expected the number of examples in the training set have an impact on the resulting accuracies.
Usage of the largest training set $10 \times \textsc{All}$ is more accurate than usage of the smallest set $1/2 \times 1/2$.
In the $10 \times \textsc{All}$ dataset the amount of examples per class in the training set vary between 40 and 67 which is why a splitting of $10\times 40$ is only slightly worse, even though the errors of each results are too large to allow a significant statement.

Since our optimums and the ones of Sulc and Matas \cite{Sulc:2015} coincide within their errors we can not state that our results are significantly better on the Flavia dataset, but we appreciably outperform the state-of-the-art value  on the Foliage dataset.
However, we definitely achieve significantly improved classification accuracies of CNNs on these datasets compared to the proposed settings by Zhang et al. \cite{Zhang:2015}.

\subsection{Error discussion}

\begin{figure}[htb]
	\centering
	\includegraphics[width=\linewidth]{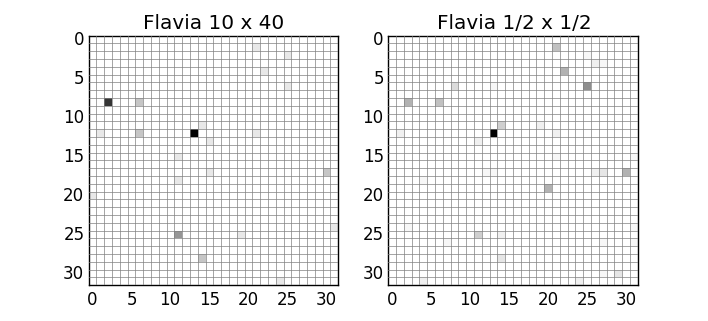}
	\caption{This figure shows the misclassifications on the Flavia dataset for the $10\times 40$ and $1/2 \times 1/2$ validation methods. The blacker a matrix entry the more often occurred an error of the notion: ``a member of class $y$ was misclassified as class $x$''.}
	\label{fig:errors}
	\includegraphics[width=0.5\linewidth]{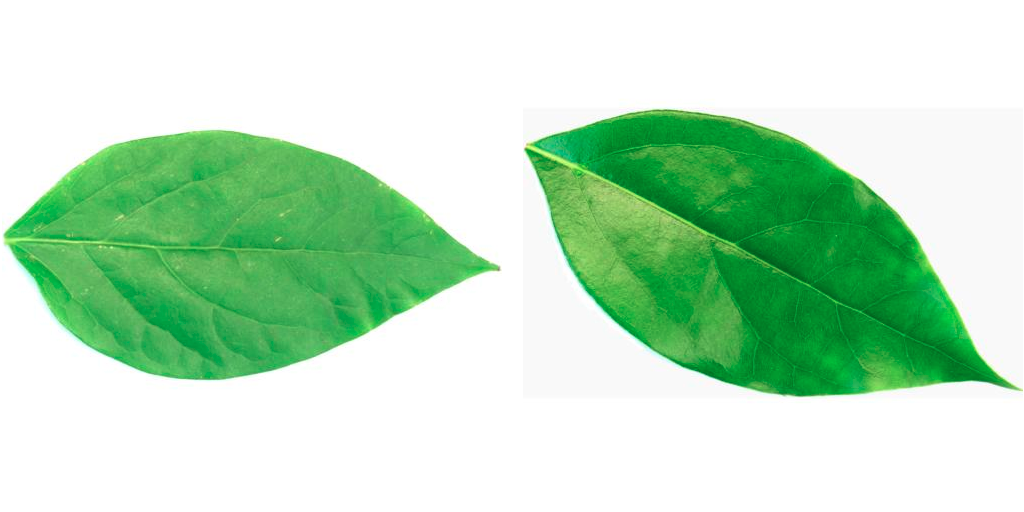}
	\caption{The left leaf is a member of the species Chimonanthus (wintersweet)  whereas the right leaf is Cinnamomum camphora (camphor tree).}
	\label{fig:flavia_12_13}
\end{figure}

To examine the errors of the model we show the misclassifications on the Flavia dataset in Fig. \ref{fig:errors}.
Therefor we sum up all wrong predictions of all models ($T_0$, $T_R$, $T_F$) of a single splitting in a matrix whose rows indicate the correct labels and whose columns show the misclassifications.
Each single plot is normalized to its maximum that is drawn black.

At first, the distribution of the black boxes attract attention.
They are not randomly distributed but instead ordered in rows or columns, i.e. if there exists a box at a certain position it is likely that there is another box in the same row or column.
Moreover, the shape is not symmetrical.
There are only some examples where pairs of connected misclassifications exist, e.g. $(6, 8)$ and $(10, 6)$ in Flavia $10\times \textsc{All}$.
The interpretation is that two classes are most commonly not similar to each other, but that similarity of one leaf to one or more other leaves is a unidirectional quantity.

If one looks deeper into single nodes a noticeable commonality is the matrix entry $(13, 12)$ i.e. 12 was misclassified as 13, that is dominant in all three plots.
Some exemplary members of these species are shown in Fig. \ref{fig:flavia_12_13}.
Interestingly this error is not symmetrical, i.e. members of class 13 are identified correctly whereas members of class 12 are more likely to be misclassified.

Furthermore, there are several other classes that are tough for our network to identify. 
On the one hand these are $(2,8)$ and $(6,8)$ that means that members of class 8 are more likely to be misclassified as 2 or 6, and on the other hand $(30, 17)$, $(12, 25)$, and $(14, 28)$.

%===============================================
\section{Conclusions}
%===============================================

Our proposed CNN was trained by applying the presented variety of improvements to overcome the small amount of training examples in the Flavia and Foliage dataset. It yields results with state-of-the-art performance of above 99\% for leaf identification on the Flavia and Foliage dataset.
The presented method is in our best knowledge currently the best CNN approach on this task, mainly by using transfer learning and data augmentation.
Thereby we combine data augmentation and training into one step to create a model that can be adopted easily to other applications.
Even compared to methods based on handcrafted features we showed that supervised training of CNNs yield slightly better results.

As a next step we will apply the presented methods on the larger MEW dataset \cite{Novotny:2013} that contains 153 species and at least 50 per class.
We expect competitive accuracies that are definitively worse than the results on the Flavia dataset, because the number of instances per class is approximately the same but the total number of classes is almost five times as big.

However, our main goal is to apply the described methods to cross dataset validation.
Comparable to \cite{Reul:2016} we want to study how a trained CNN applies to a real world scenario, i.e. classifying leaves that were collected completely independent of the test set and therefore differ quite a lot due influences of the temperature, rainfall, solar irradiation or seasons.
Similar to \cite{Reul:2016} we expect that the applications of convolutional neural networks that are highly optimized on the trained dataset will fail on cross dataset validations especially due to the usage of colors, which is why experiments utilizing gray-scale images are required to achieve improved results.
Another approach would be to generate augmentations allowing color changes in a defined way so that the network learns possible different colors of leaves.

%==========================================================================
\appendix
%==========================================================================

%==========================================================================
\bibliography{paper}
\bibliographystyle{splncs03}

\end{document}